\pdfoutput=1
\documentclass[11pt]{article}

\usepackage[final]{acl}

\usepackage[utf8]{inputenc}
\usepackage[T1]{fontenc}
\usepackage{newtxtext}
\usepackage{newtxmath}
\usepackage{inconsolata}
\usepackage{microtype}

\usepackage{graphicx}
\usepackage{booktabs} 
\usepackage{tabularx}
\usepackage{multirow}
\usepackage{siunitx}
\usepackage{pbalance}

\title{%
    Dialogue Is Not Enough to Make a Communicative BabyLM\\
    (But Neither Is Developmentally Inspired Reinforcement Learning)
}

\author{
    \textbf{Francesca Padovani\textsuperscript{1}\thanks{These authors contributed equally.}} \:
    \textbf{Bastian Bunzeck\textsuperscript{2}\footnotemark[1]}\:
    \textbf{Manar Ali\textsuperscript{2}} \:
    \textbf{Omar Momen\textsuperscript{2}} \\
    \textbf{Arianna Bisazza\textsuperscript{1}} \:
    \textbf{Hendrik Buschmeier\textsuperscript{2}} \:
    \textbf{Sina Zarrieß\textsuperscript{2}}
\\
    \textsuperscript{1}{Center for Language and Cognition (CLCG), University of Groningen}\\
    \textsuperscript{2}{CRC 1646 -- Linguistic Creativity in Communication, Bielefeld University}
 \\
    \texttt{f.padovani@rug.nl} \:
    \texttt{bastian.bunzeck@uni-bielefeld.de}
 }

\begin{document}
\maketitle

\begin{abstract}
    We investigate whether pre-training exclusively on dialogue data results in formally and functionally apt small language models. Based on this pre-trained \texttt{llamalogue} model, we employ a variety of fine-tuning strategies to enforce ``more communicative'' text generations by our models. Although our models underperform on most standard BabyLM benchmarks, they excel at dialogue continuation prediction in a minimal pair setting. While PPO fine-tuning has mixed to adversarial effects on our models, DPO fine-tuning further improves their performance on our custom dialogue benchmark. 
\end{abstract}

\section{Introduction}
\label{sec:introduction}

Large language models are capable of generating language with almost human-like fluency. To do so, however, they need unfathomable amounts of textual input as training data. In comparison, humans are highly sample-efficient learners and develop a full-fledged linguistic system from input that is orders of magnitude smaller. In the past, this sample efficiency has mostly been attributed to genetically pre-endowed priors \citep{chomsky1986knowledge, berwick2011poverty}. More recently, the quantitative, usage-based turn in linguistics has focused on the importance of language use, interaction and grounding in the real world and more domain-general cognitive mechanisms for language learning \citep{tomasello2003constructing, tomasello2005formalities, behrens2021constructivist}. Crucially, language is primarily a tool for communication \citep{fedorenko2024languagea, levinson2025interaction}, and therefore all acquisition processes must be conceptualized accordingly. 

Lately, the BabyLM paradigm has emerged as a novel way of testing claims of learnability with little data, small language models and linguistically inspired evaluation tasks \citep{warstadt2023findings, hu2024findings, charpentier2025babylm}. Although highly optimized models are indeed able to capture linguistic structure very accurately (e.g., \citealp{charpentier2023not, tastet2024babyllama2}), they are still trained on a wider variety of input registers than the main input modality of children, namely child-directed speech in dialogue. Observed in isolation, child-directed speech does differ tremendously from other input modalities, featuring many fragments, more questions and less canonical SV(X) sentences \citep{cameron-faulkner2003construction, bunzeck2025richness}. Despite \citet{huebner2021babyberta} finding it to be conducive pretraining data for simplified benchmarks, more recent work has shown that its effects can be described as \textit{mixed} at best \citep{padovani2025childdirected, bunzeck2025construction}.

One possible explanation for this discrepancy is that autoregressive language models, trained on a next-token prediction task, do not model the communicative aspects that are seen as crucial for language acquisition and underlie the fragmented nature of child-caregiver dialogue. Common data pre-processing protocols for BabyLMs split child-caregiver dialogues into isolated sentences, which effectively removes communicative context that is available and central for human learners. Therefore, we conceptualize the task of training a BabyLM differently: We train a small, autoregressive model\footnote{Given its training on dialogue data only, throughout this paper we refer to the base model as \texttt{llamalogue}. All models and datasets can be found in this \href{https://huggingface.co/collections/CLAUSE-Bielefeld/communicative-babylm-68dfb2d876e9ef11fb7e6751}{Huggingface collection}.} on dialogue triplets extracted from CHILDES \citep{macwhinney2000childes}. As such, our model is not a model of the learner \textit{per se}, but of the interaction and communication underlying the language learning process. Additionally, we apply different reinforcement learning paradigms to our model to make the ‘child’ component of the dialogue system more fluent and contextually appropriate when interacting with a ‘caregiver’ dialogue partner. In sum, we test the following ideas through this process:
(i) How does a BabyLM trained only on child--caregiver dialogue  perform? And (ii) Are there ways of teaching BabyLMs to be more communicative speakers via interaction and communication? 

We find that (i) our base model pre-trained exclusively on child--caregiver dialogues maintains above-chance accuracy on formal linguistic competence, while achieving higher accuracy in predicting realistic communicative turns than a baseline autoregressive model. Moreover, (ii) directly aligning preferred child responses to caregiver utterances through DPO proves more effective than interactively fine-tuning the policy via PPO with a reward function, especially when evaluated on dialogue minimal pairs. However, none of these fine-tuning techniques improves performance on more formal benchmarks.

\section{Related work}
\label{sec:related-work}

\paragraph{Learning exclusively from CDS}

While the standard English BabyLM corpus consists of approximately 30\% child-directed speech, ample work exists on pretraining LMs from scratch on 100\% child-directed speech (CDS). In a seminal paper, \citet{huebner2021babyberta} showed that a small 5M-parameter BabyBERTa model, trained on 5M lexical tokens of child-directed speech, shows the same accuracy on Zorro \cite[vocabulary-limited minimal pair tasks;][]{huebner2021babyberta} as the RoBERTa-base model with 125M parameters and trained on 30B words. Similar results are presented by \citet{feng2024childdirected}, who show that autoregressive models trained on CDS alone perform only slightly worse on Zorro than comparable architectures trained on Wikipedia data, synthetic data, or the BabyLM corpus. However, their CDS models underperform other models tremendously on semantic similarity benchmarks. Negative results are also reported by \citet{yedetore2023how}, who show that autoregressive models trained on CHILDES data fail to reliably acquire hierarchical generalizations in question formation from declaratives, and rather prefer incorrect linear generalizations. 

Expanding the CDS-only training paradigm to more languages than English, \citet{salhan2024less} find that developmentally-inspired curriculum learning strategies during pretraining improve scores on syntactic minimal pairs for models trained on English, French, German, Chinese, or Japanese CDS, outperforming models trained on Wikipedia data by over 10\%. Conversely, \citet{padovani2025childdirected} report less positive results. For many syntactic minimal pair benchmarks, their CDS models underperform in comparison to Wikipedia-trained models across different languages (English, German, French). Finally, \citet{bunzeck2025construction} approximate German CDS on the level of utterance-level construction distributions. They also find that models trained on it are generally inferior to models trained on comparable Project Gutenberg data when evaluated on syntactic benchmarks, although the CDS models show moderate improvements on some word-level benchmarks. 

In sum, it can therefore be said that pre-training on CDS is only conducive to language model performance for highly specific benchmarks like Zorro (although results are inconsistent across studies) or in more specific training regimens like curriculum learning.

\paragraph{Cognitively/developmentally plausible RL}

Despite reinforcement learning, especially in the form of RLHF \citep{ouyang2022training}, being an integral part of modern language modeling practices, it has only very recently begun to get adopted in cognitively inspired modeling.
\citet{zhao2023babystories} improve their small models trained on BabyLM data by constructing a RLHF dataset from human-annotated story continuations generated with regular GPT-2 and then reinforcing these storytelling capabilities of their models. While it does not improve performance on zeroshot benchmarks, it makes their models better base models for fine-tuning tasks. 

In a more developmentally inspired fashion, \citet{ma2025babysit} generate text continuations from a student GPT-2 model and compare these to an already further trained teacher model. A reward signal is then generated from the model's estimated ‘age’ (\textit{viz.} training steps), based on its continuations and the teacher continuation. This interactive learning is then interleaved with regular causal language modeling. Their interactive model outperforms regular autoregressive models on word acquisition, quantified as average surprisal for a set of test sentences.

\citet{stopler2025developmentally} introduce a training regime inspired by emergent communication research that again includes two language models: a speaker/child language model, and a listener/caregiver language model. In their setup, the speaker model has to summarize a passage, and the listener model has to answer a question solely based on the summary provided by the speaker. If the listener model (whose weights are frozen) answers correctly, a reward signal is used to update the speaker model. Although their reinforcement strategy changes speaker behavior, it does not improve performance on linguistic benchmarks.

Finally, \citet{nikolaus2025modeling} base their reward signal on an annotated dataset of CHILDES data with clarification requests in parental utterances, which often trigger children to use more ``grammatical'' language. For each utterance produced by their child language model, they predict if it would possibly beget such a clarification request, and reward productions that do not. They show that this process improves their models on the reinforcement goal of producing less ``ungrammatical'' utterances, but has mixed to no effects on grammar benchmarks like BLiMP \citep{warstadt2020blimp} and Zorro.

\section{Methodology}
\label{sec:methodology}

\subsection{Pretraining}
\label{sec:pretraining}

\paragraph{Data}

Our models are trained on dialogue data from the English CHILDES section. In a first preprocessing step, we clean the transcripts from CHILDES quite heavily by removing all extra- and paralinguistic information. Furthermore, we replace all unintelligible or otherwise incomplete utterances, for which annotations as to the intended word are available, with these intended words. Finally, we split all utterances that contain explicitly annotated pauses, as there is no clear distinction between such pauses and utterance boundaries marked by regular line breaks.

From these cleaned dialogues, we extract all utterance triplets (three consecutive turns) where at least two different speakers are involved. Furthermore, we enforce the triplets to contain at least five lexical words. This excludes triplets that only contain repetitions of single words or are otherwise light on lexical content. We leave the speaker tags in the data. A typical line from our data might therefore look as follows:
\begin{verbatim}
*CHI: all gone .
*MOT: where's the kitty ?
*CHI: all gone .
\end{verbatim}
By using dialogue data only, we assume that the autoregressive pretraining process pushes our BabyLM to model contingent structure (responses depend on previous turns), learn turn-level coherence, and acquire some knowledge about implicit expectations in communication, e.g., that questions beget a response.

\paragraph{Base model} 

We train a small 135M-parameter Llama model \citep{touvron2023llama} on 10M lexical tokens from the aforementioned set of dialogue triplets. Our model features 16 layers, 16 attention heads and a hidden/intermediate size of \num{1024}. We train the model for 10 epochs. As we found approximately 60k different lexical types in our data, we opt for a small vocabulary size to not store too many of these types holistically. We fit a BPE tokenizer on the training data to include 8k tokens. Crucially, we fit the tokenizer on the actual transcriptions only, not on the speaker tags. The speaker tags are added as additional tokens afterwards. In sum, with the inclusion of all speaker tags, this results in a vocabulary size of \num{8465}.

\subsection{DPO fine-tuning}

As a first attempt to further align \texttt{llamalogue} with child-like, communicatively appropriate behavior, we employ Direct Preference Optimization \citep[DPO;][]{Rafailov2023}, which is a preference-based training method that directly optimizes the model to prefer certain continuations over others. In our case, this procedure is supposed to guide our model to favor contextually appropriate utterances over random ones.

\paragraph{Naturalistic data}
\label{sec:naturalistic_data}

As fine-tuning data, we construct a dataset of minimal dialogue pairs derived from another set of triplets not seen during pretraining and not used for validation. From these, we extract naturally occurring caregiver--child exchanges and derive  contrastive, incorrect variants by replacing the child utterance with a randomly sampled one. 
To systematically control for confounds, we focus on minimal pairs that are matched in length (by number of words or subword tokens) and filter out pairs where the child utterance repeats words from the caregiver utterance, resulting in approximately \num{26000} pairs. For DPO training, we select the word-matched minimal pairs, subsampling \num{18000} examples for training, with the remaining \num{8000} examples held out for evaluation. Overall, the fine-tuning phase was conducted on a total of \num{245480} tokens.

\paragraph{Synthetic data}

In addition to the real data, we generate a synthetic DPO dataset to probe the benefits of model-guided preference generation. Here, the caregiver's utterance is used as a prompt to Llama-3.2-3B \citep{touvron2023llamaa}, which generates a plausible child response. Incorrect alternatives are again randomly sampled from the original dataset. Here, we do not control for matched length, as the exact number of generated words is not easy to control. The child continuation is generated through an instructive prompt (cf. Table~\ref{tab:prompt_generation}) designed to facilitate short and natural completions. In total, the synthetic training data is composed of \num{245480} tokens.

\begin{table}
\centering
\small
\begin{tabular}{p{0.93\columnwidth}}
\toprule
    You are a young child having a conversation with your mother. \\
    When your mother says something, you should answer as a typical and natural-sounding child. Do NOT repeat her words. Instead, give a new, relevant answer that shows understanding. \\
    Keep it short and child-like.\\
    \\
    *MOT: I think they just throw it on the side . \\
    
    *CHI: \\
\bottomrule
\end{tabular}
\caption{Zero-shot prompt to Llama-3.2-3B.}
\label{tab:prompt_generation}
\end{table}

The two fine-tuning datasets are available in our Huggingface collection. Representative examples from both datasets are provided in Appendix~\ref{sec:sample_dpo_datasets}. 

\begin{figure*}[t]
  \centering
  \begin{minipage}[t]{0.5\textwidth}
    \centering
    \includegraphics[width=\linewidth]{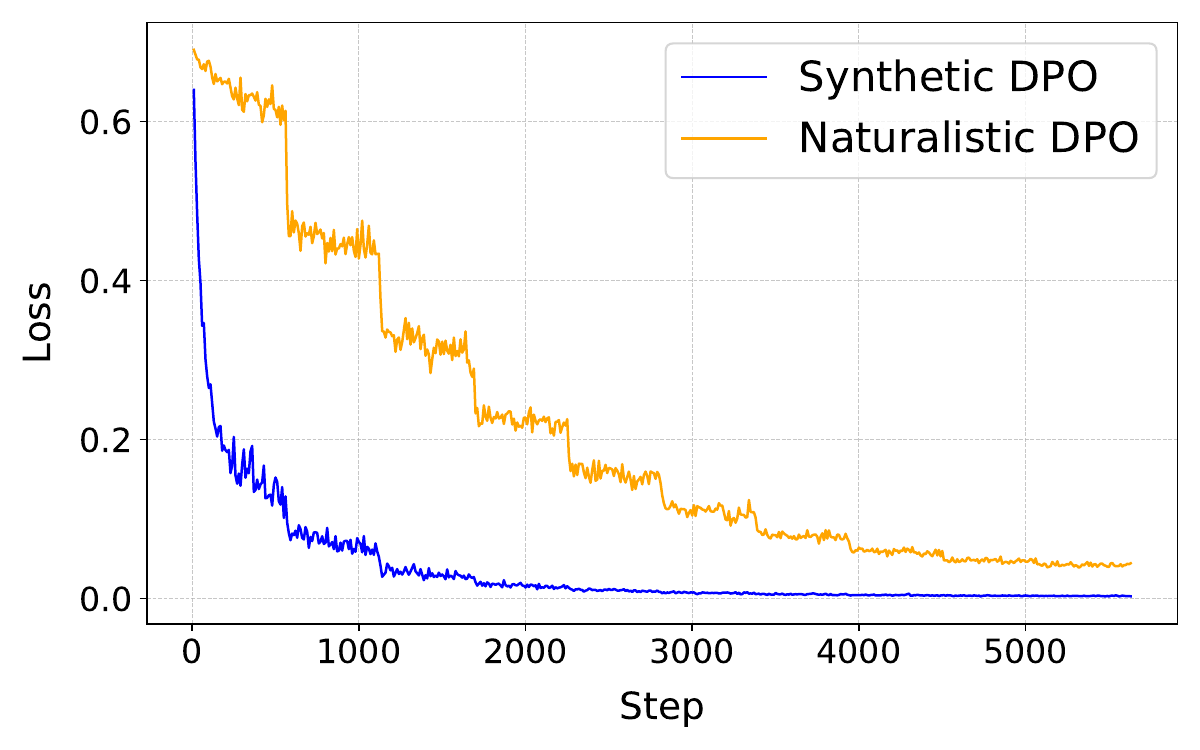}
  \end{minipage}\hfill
  \begin{minipage}[t]{0.5\textwidth}
    \centering
    \includegraphics[width=\linewidth]{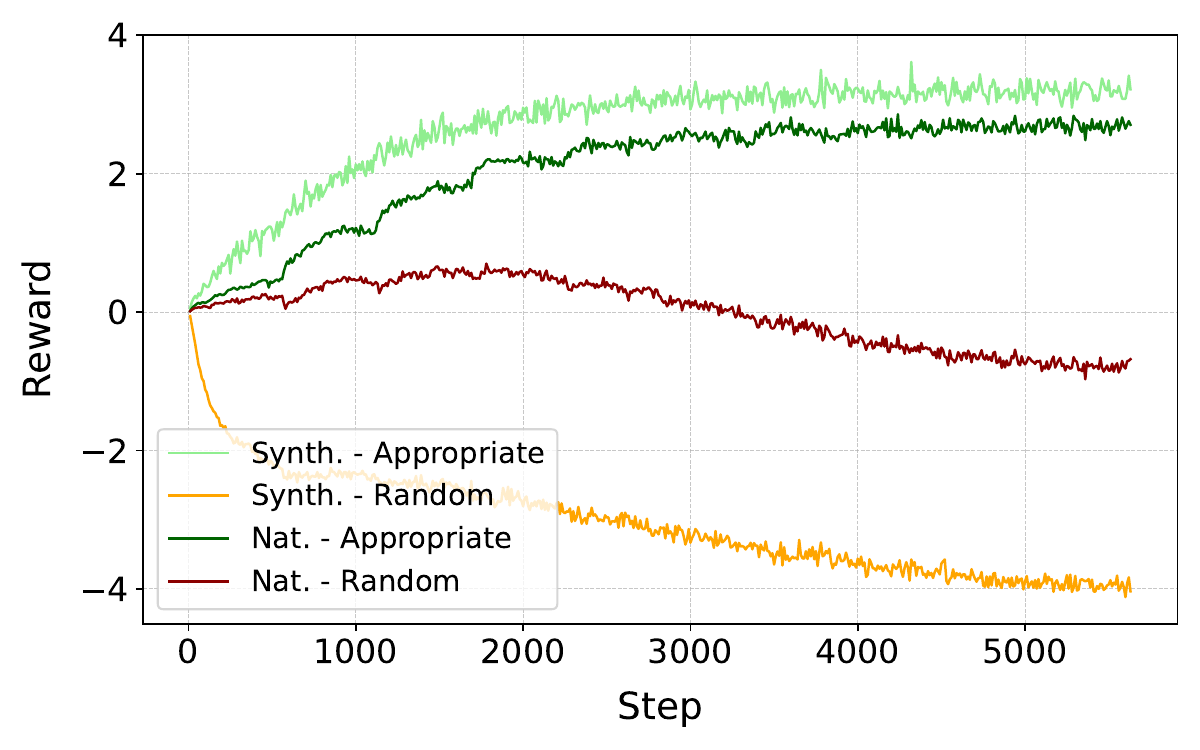}
  \end{minipage}
  \caption{Training loss (left) and reward trends (appropriate vs. random) during training (right) for both DPO models.}
  \label{fig:dpo_rewards_loss}
\end{figure*}

\medskip
We perform one 10-epoch DPO fine-tuning run with \texttt{llamalogue} on each dataset with \texttt{trl}  \citep{vonwerra2022trl}. A learning rate of $5 \times 10^{-6}$ is used, with a per-device batch size of 4, and 4-step gradient accumulation, resulting in an overall batch size of 16. Figure~\ref{fig:dpo_rewards_loss} shows the two loss and reward trends for the appropriate and random sentences of the fine-tuning datasets. 

\subsection{PPO fine-tuning}
\label{sec:ppo_intro}

To steer the communicative behaviour of \texttt{lla\-malogue} more indirectly, we also fine-tune it using Proximal Policy Optimization \citep[PPO;][]{Schulman2017}. To implement the notion of `effective communication' for PPO, we needed to substantially simplify it. Developmental research has extensively characterized learning as involving a dynamic exploration--exploitation trade-off \citep{kim2024, gopnik2020, nussenbaum2019}, in which children alternate between experimenting with novel behaviors (i.e., linguistic forms) and leveraging familiar patterns. However, operationalizing this \emph{sweet spot} between exploration and exploitation as a computational reward function is inherently difficult. To formalize what constitutes a ``successful'' communicative turn, we explored a range of reward functions reflecting different aspects of communication: a BLEU-based reward, a semantic similarity reward, a quality score derived from an LLM, and an uncertainty-based reward measuring LLM confidence in processing child responses.

Our PPO pipeline requires real caregiver prompts as input for both \texttt{llamalogue} and a teacher LLM emulating a ``good, communicative baby'', therefore we extract caregiver utterances (minimum four tokens length) from unused segments of the pre-processed CHILDES dialogue triplets. We then prompt a teacher LLM, as before a Llama-3.2-3B \citep{touvron2023llama}, with these utterances, asking it to generate candidate responses simulating a short child-like answer that shows understanding of the caregiver utterance\footnote{%
    Examples can be found in Appendix~\ref{sec:ppo_ref_sent}.
}. The prompt is the same as the one used for generating the DPO datasets (Table~\ref{tab:prompt_generation}). The reward functions are then computed by comparing these teacher-generated responses to the output produced by \texttt{llamalogue} in response to the same utterance. Calculating the reward as an average over 10 generated responses proved to be noisy due to their variability, so we ultimately based the reward on the comparison between the model's output and the one single response generated by the teacher LLM.

\paragraph{1-gram BLEU Reward}

The BLEU-based metric \citep{papineni2002} captures surface-level lexical similarity. Specifically, we compute a smoothed unigram BLEU score (BLEU-1) between \texttt{llamalogue}'s response and the teacher LLM's reference answer with \texttt{nltk}. We apply smoothing to avoid zero scores. The resulting reward values range from 0 to 1.

\paragraph{Semantic Similarity Reward}

As a complementary approach to lexical overlap, we also implement a semantic similarity reward to promote contextually appropriate, meaningful responses. Specifically, we use the \texttt{all-MiniLM-L6-v2} model from \texttt{SentenceTransformers} \citep{reimers-2019-sentence-bert} to compute the cosine similarity between the BabyLM’s response and the reference utterance generated by the teacher LLM. This similarity score, ranging from 0 to 1, encourages outputs that align semantically with high-quality examples.

\paragraph{LLM-generated Reward}

To further explore reward signals grounded in communicative quality, we prompt an LLM to directly assess \texttt{llamalogue}'s responses. Given a caregiver utterance and the generated child continuation of \texttt{llamalogue}, the LLM is instructed to assign a numerical quality score (from 0 to 5) based on contextual appropriateness and fluency (see Table~\ref{tab:olmo-prompt}). After experimenting with various models, including Llama-3.2-3B and Nemotron-Research-Reasoning-Qwen-1.5B \citep{liu2025}, we selected OLMo-2-1124-7B-Instruct \citep{OLMo-etal-2025-features}. This choice was motivated by the fact that OLMo consistently adhered to the requested output schema and avoided formatting anomalies that hindered automated reward extraction. The scalar score returned by OLMo is then used as the reward signal during each PPO step.

\begin{table}
\centering
\small
\begin{tabular}{p{0.93\columnwidth}}
\toprule
    \texttt{<|system|>} \\ 
    You are presented with a dialogue between a mother (MOT) and a child (CHI). \\
    Please rate how contextually appropriate and fluent the child's response is, on a scale from 0 (completely unfitting) to 5 (perfectly fine answer). If CHI answer is too short rate it low. \\
    \texttt{<|end|>} \\
    \texttt{<|user|>} \\
    MOT: It's like in the grocery store when go shopping . \\ 
    CHI: Mom, please let me choose the food for myself. \\
    \texttt{<|end|>} \\
    \texttt{<|assistant|>} \\
\bottomrule
\end{tabular}
\caption{Zero-shot prompt to OLMo.}
\label{tab:olmo-prompt}
\end{table}

\paragraph{Teacher Confidence-based Reward}

To incorporate a measure of uncertainty into the reward signal, we implement a confidence-based metric: For each caregiver utterance $x$, we precompute the log-probabilities $\{\ell_i\}_{i=1}^{10}$ assigned by the frozen Llama-3.2-3B to the set of 10 reference child responses generated by that same model $\{y_i\}_{i=1}^{10}$, as explained in Section~\ref{sec:ppo_intro}. During fine-tuning, the BabyLM’s generated response $\tilde y$ is scored using the same teacher to obtain 
\begin{equation*}
    \ell_{\text{baby}} \;=\; \log P_{\text{teacher}}(\tilde y \mid x)
\end{equation*}
Then we compute the normalized rank \[
\mathrm{rank}(x,\tilde y)\;=\;\frac{1}{10}\sum_{i=1}^{10}\mathbf{1}\!\left\{\ell_i \le \ell_{\text{baby}}\right\}\in[0,1],
\] and linearly map it to a PPO reward \[
r(x,\tilde y)\;=\;2\,\mathrm{rank}(x,\tilde y)-1\;\in[-1,1].
\]
This signal favors BabyLM outputs that the teacher assigned high likelihood and potentially bias the model towards more grammatical and distributionally expected utterances.

\begin{table*}[ht]
\centering
\footnotesize
\begin{tabularx}{\textwidth}{clcccccccc}
\toprule
 &  &  & \multicolumn{2}{c}{\textbf{DPO}} & \multicolumn{4}{c}{\textbf{PPO}} &  \\ 
 \cmidrule(lr){4-5}\cmidrule(lr){6-9}
 & \textbf{Task} & \textbf{\texttt{llamalogue}} & \textbf{Natural.} & \textbf{Synth.} & \textbf{Bleu} & \textbf{SemSim} & \textbf{LLM Score} & \textbf{Conf.} & \textbf{Baseline} \\ \midrule
\multirow{10}{*}{\rotatebox{90}{\shortstack{Zero-shot\\ (Baby LM)}}} 
 & BLiMP & 56.05 & 55.64 & 55.51 & 55.14 & \textbf{56.36} & 55.31 & 55.10 & 72.16 \\
 & BLiMP suppl. & 51.06 & 49.97 & \textbf{51.67} & 51.33 & 51.48 & 50.58 & 49.45 & 61.22 \\
 & COMPS & 51.62 & 51.51 & \textbf{51.63} & 50.66 & 51.58 & 51.25 & 51.59 & --- \\ 
 & Entity tracking & 30.66 & 32.66 & 31.29 & 16.20 & 34.64 & \textbf{36.03} & 34.05 & 28.06 \\
 & EWoK & 50.19 & 50.12 & \textbf{50.82} & 49.65 & 49.62 & 50.12 & 50.81 & 51.92 \\
 & Read. (eye track.) & \textbf{3.88} & 3.57 & 1.16 & 3.43 & 2.85 & 3.73 & 3.35 & 9.08 \\
 & Read. (self-paced) & 1.43 & 1.35 & 0.44 & \textbf{1.99} & 1.04 & 1.30 & 1.14 &  3.5 \\
 & Wug adj. & 0.45 & 0.52 & 0.16 & 0.13 & 0.01 & \textbf{0.55} & 0.41 & 38.5 \\
 & Wug past & -0.03 & -0.01 & -0.05 & -0.15 & -0.18 & -0.01 & -0.19 & --- \\
 & AoA & -79.6  & 0  & 0 & -80.1 &  0  & -76.6 & -78.7 & --- \\
 \midrule
FT & (Super)GLUE & 51.82 &	51.72 &	51.77 &	51.12 &	52.10 &	51.69 &	\textbf{51.92} & 67.91 \\ \midrule
\multirow{4}{*}{\rotatebox{90}{\shortstack{Zero-shot\\ (Add’l)}}} 
 & Lexical decision & 40.3 & 40.5 & \textbf{41.3} & 40.7 & 39.7 & 40.2 & 40.8 & 57.2 \\
 & Zorro & \textbf{65.5} & 64.8 & 62.7 & 62.5 & 64.7 & 65.2 & 63.7 & 77.7 \\
 & Dia. MP (Words) & 64.3 & \textbf{68.4} & 64.9 & 62 & 61.1 & 60.6 & 63.7 & 58.1 \\ 
 & Dia. MP (Tokens) & 63.8 & \textbf{67.6} & 64.3 & 61 & 63.6 & 62.5 & 62.4 & 57.9 \\ \bottomrule 
\end{tabularx}
\caption{Full results for pre-trained and fine-tuned (FT) models. For each task, the best-performing model among those we pre-trained and fine-tuned (excluding the baseline) is shown in bold.}
\label{tab:full-results}
\end{table*}

\paragraph{Training configuration}

In our experimental trials we rely on the default PPO training parameters provided by the \texttt{trl} library for all fine-tuned models, with the exception of the one trained using the Teacher Confidence-based reward. This reward caused higher variance in the reward values, making the KL control more sensitive. Therefore we set the KL penalty mode to \texttt{abs}, a lower learning rate of $5 \times 10^{-6}$ and a small initial KL coefficient of $0.02$ to weaken the penalty for policy updates in the early stage of training. 

Moreover, the fine-tuning processes based on the first three PPO strategies employed a larger portion of the training data, as caregiver utterances inputs, from our original pre-processed set (\num{220000}) compared to the model fine-tuned using the Teacher Confidence-based reward (\num{150000}). In the latter case, we observed that a lower number of training steps was sufficient to achieve a consistent, significant improvement in the reward.
We fine-tune for 3 epochs, with each epoch featuring \num{13750} steps for the first three PPO strategies and \num{8645} steps for the Teacher Confidence-based reward.
In terms of token usage, for the first three PPO strategies we estimate a total of \num{3009104} tokens, obtained by summing the tokens from the prompts provided to \texttt{llamalogue} and the single ground-truth response generated by the teacher LLM. For the Teacher Confidence-based reward strategy, where ten teacher responses were used for confidence estimation, the total amounts to \num{9903146} tokens. In both cases, the overall token count remains well below the 100M-token limit specified by the BabyLM Challenge for the interaction track.\footnote{%
    The code for DPO and PPO experiments can be found at these two Github repositories: \url{https://github.com/fpadovani/communicative_baby_ppo} and \url{https://github.com/fpadovani/communicative_baby_dpo}
}

\subsection{Evaluation}
\label{sec:evaluation}

\paragraph{Standard benchmarks}

For evaluation purposes, we rely on the BabyLM evaluation pipeline \citep{charpentier2025babylm}. As zero-shot evaluation, it includes minimal pairs tasks on the syntactic level (BLiMP, \citealp{warstadt2020blimp}) and on the semantic/world knowledge level (COMPS, \citealp{misra2023comps}; EWoK, \citealp{ivanova2024elements}; entity tracking, \citealp{kim2023entity}). Additionally, in further tasks, model probabilities/surprisal values are correlated with word-level age of acquisition \citep{chang2022word}, 
cloze probabilities \citep{devarda2023cloze}, and preferences in morphological inflection for ‘wug’ words \citep{hofmann2025derivational}. Finally, the models are also evaluated through fine-tuning on a selection of tasks from GLUE and SuperGLUE \citep{wang2018glue, wang2019superglue}.

\paragraph{Custom benchmarks}

To evaluate the models in a more holistic way, we include three additional minimal pair benchmarks. We (i) create a dialogue minimal pair set. As already described in Section ~\ref{sec:naturalistic_data}, positive examples are created by simply matching parental utterances with children's answers, negative examples are sampled by matching the same parental utterances with unrelated child utterances. With this dataset, we aim to not only test the formal language skills of our models (as the BabyLM evaluations already do), but also their functional skills \citep{mahowald2024dissociating}. Furthermore, we include (ii) Zorro \citep{huebner2021babyberta}, a reduced version of BLiMP with a vocabulary restricted to words that occur in CHILDES, and (iii) the lexical decision dataset by \citet{bunzeck2025subword}, which contains word-level minimal pairs of words and non-words (e.g., \emph{sending} and \emph{monding}) as benchmarks that should be more tuned to the linguistic register found in our pretraining data.

\section{Results}
\label{sec:results}

\subsection{Base model evaluation}

We evaluate our base model after being trained for 10 epochs. 
We compare \texttt{llama\-logue} to the baseline model \texttt{babylm-interaction-baseline-simpo}\footnote{%
    \url{https://huggingface.co/BabyLM-community/babylm-interaction-baseline-simpo}
} provided by the BabyLM organizers for the \emph{interaction} track. Our model performs worse than this baseline model in almost every BabyLM evaluation task, except entity tracking (cf. Table~\ref{tab:full-results}). In comparison to other models submitted to the \emph{strict-small} track, our model performs particularly worse on BLiMP and AoA prediction, whereas scores for EWoK, COMPS, (Super)GLUE or the different wug tests are undercut by several other submissions. Therefore, \texttt{llama\-logue} is not a generally bad language model, but its pretraining peculiarities have a non-straightforward effect on performance.

With respect to the custom benchmarks the results are more nuanced. For example, on the \emph{dialogue minimal pairs} task, which aligns closely with the pre-training goal of \texttt{llamalogue}, it exhibits a clear advantage over the baseline comparison (63--64\% vs. 57--58\%). Our model also achieves a reasonable accuracy of 65.5\% on \emph{Zorro}. Nevertheless, it is clearly outperformed by the interactive baseline (77.7\%) which was trained on the full BabyLM data. Our base model also falls behind the interactive baseline on the \textit{lexical decision} task, performing quite far (40.3\%) below chance level. 

\begin{figure*}
  \centering
  \begin{minipage}[t]{0.24\textwidth}
    \centering
    \includegraphics[width=\linewidth]{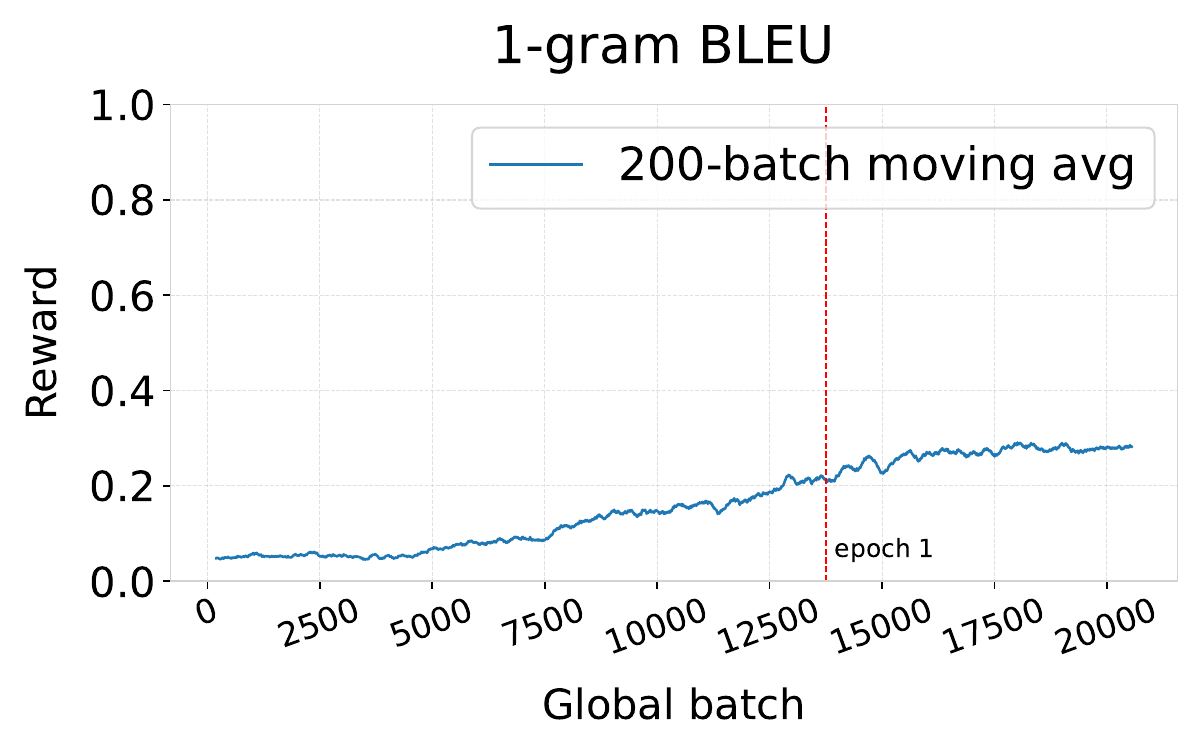}
    \label{fig:bleu}
  \end{minipage}\hfill
  \begin{minipage}[t]{0.24\textwidth}
    \centering
    \includegraphics[width=\linewidth]{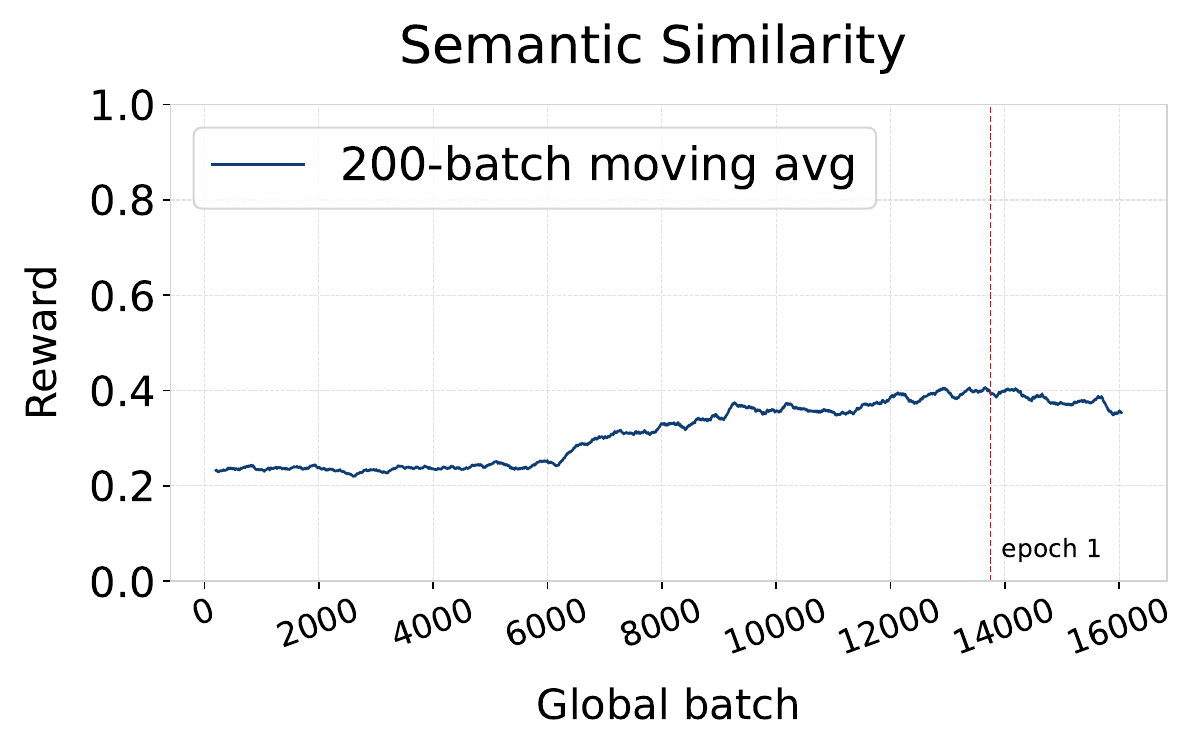}
    \label{fig:semsim}
  \end{minipage}\hfill
  \begin{minipage}[t]{0.24\textwidth}
    \centering
    \includegraphics[width=\linewidth]{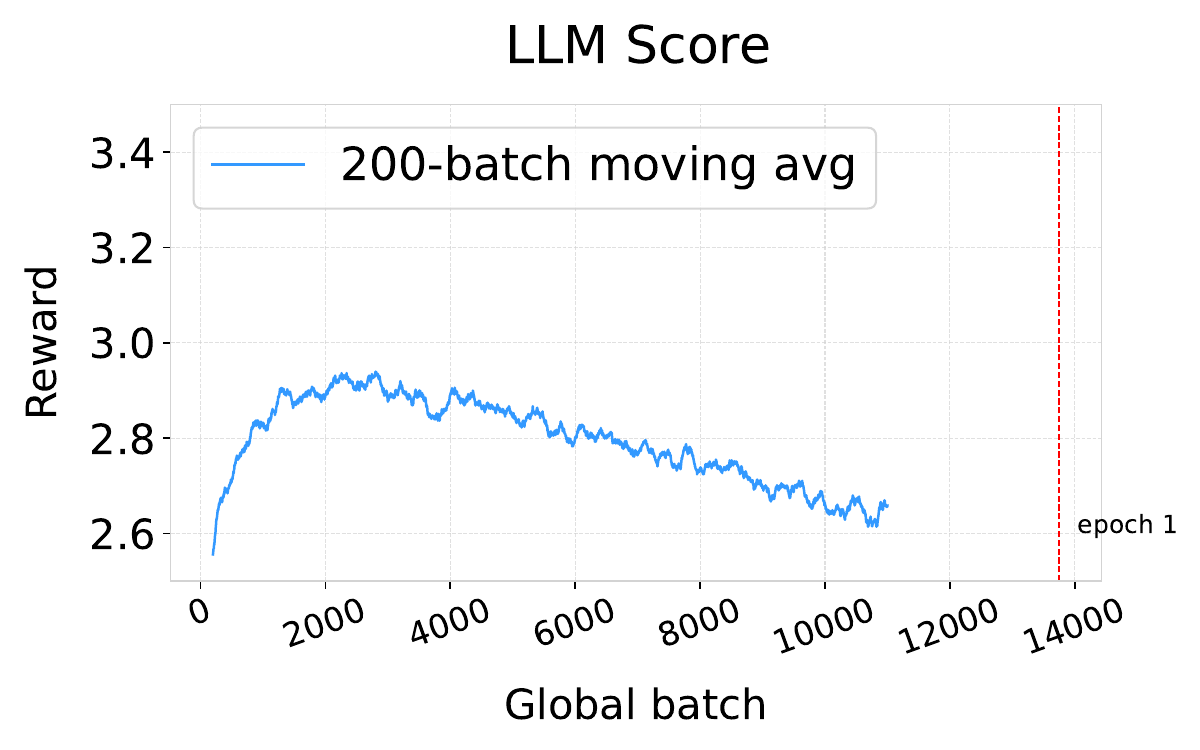}
    \label{fig:score}
  \end{minipage}\hfill
  \begin{minipage}[t]{0.24\textwidth}
    \centering
    \includegraphics[width=\linewidth]{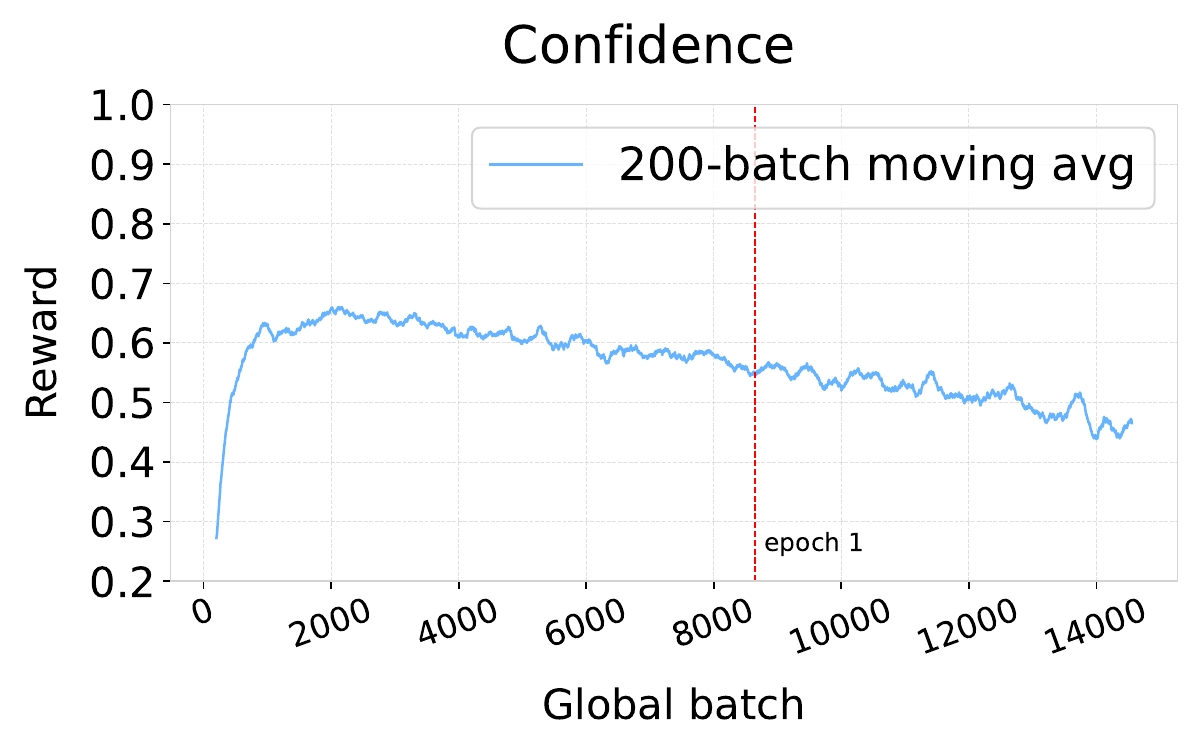}
    \label{fig:confidence}
  \end{minipage}
  \caption{Reward trends over training steps for four reward metrics: 1-gram BLEU, Semantic similarity, LLM score, and Teacher Confidence-based. Vertical line marks the end of epoch 1. For LLM score and confidence, the $y$-axis range has been restricted to enhance visibility of trends, and does not represent the full possible reward scale.}
  \label{fig:all_rewards}  
\end{figure*}

\subsection{Fine-tuned models}

\subsubsection{BabyLM evaluations}

Like the \texttt{llamalogue} base model, our fine-tuned models show overall lower performance on almost all of the zero-shot BabyLM Challenge tasks than the baseline model and the other models submitted to the interaction track. For BLiMP, all model variants score substantially below the baseline's 72.16\%, with results clustering around chance level. The highest score is achieved by the Semantic Similarity model at 56.36\%. Similar trends hold for BLiMP supplementary, where the gap to the baseline remains notable. Surprisingly, for entity tracking our models improve over the baseline of $28.06$, with the best score ($36.03$) achieved by a model fine-tuned with OLMo Score. For EWoK, scores are near chance level, in accordance with the baseline model. Reading-based tasks (eye-tracking and self-paced reading) show much lower alignment with human patterns than the baselines. 
The Wug adjective and Wug past morphological generalization tasks yield near-zero or negative correlations across all models, far from the baseline model score of $38.5$ for Wug adjective, underscoring persistent difficulty in capturing human-like morphological generalization.
For AoA, after a closer look at metric computation, we find that only very few data points (1--5 words) are considered. This is due to an unpassed condition on the parameters of the fitted sigmoid function within AoA computation in the evaluation pipeline. Limited data points lead to either a score of zero or a strong negative correlation; hence, these results can be misleading.
Overall, while entity tracking shows a modest improvement over the baseline, most linguistic and psycholinguistic tasks still reveal substantial gaps. The usefulness of our models for fine-tuning is not affected by reinforcement learning, indicated by (Super)GLUE scores that do not change drastically and also remain lower than for the baseline model (cf. also Appendix~\ref{apx:glue}).

\subsubsection{DPO reward and custom evaluations}

As shown in the right plot of Figure~\ref{fig:dpo_rewards_loss}, the reward assigned to acceptable and unacceptable utterances begins to diverge early in the fine-tuning process. This separation is particularly pronounced in the case where the acceptable sentence is artificially generated by the LLM, suggesting a stronger initial reward signal and a more stark contrast between both continuations. Interestingly, this tendency is not confirmed by performance on \emph{dialogue minimal pairs}. Although both DPO models improve upon the base model with regard to this measure, the effect of the synthetic data is rather low (increase of approximately $0.5\%$). 

In contrast, the model fine-tuned on real caregiver–child interaction data scores approximately 4\%  higher than the base model and the model fine-tuned on artificially generated child utterances. This suggests that, although LLM-generated utterances may be more grammatical and exhibit greater syntactic and lexical variety than real data found in CHILDES, the model fine-tuned on synthetic data is less apt at predicting real minimal pairs derived from genuine interactions. The natural data is clearly superior to synthetic data when trying to optimize for this task. For Zorro, the naturalistic model maintains performance comparable to \texttt{llamalogue}, whereas the synthetic model shows slightly lower accuracy.

\subsubsection{PPO reward and custom evaluations}

During PPO fine-tuning, we observe occasional instability\footnote{%
    Including abrupt drops in reward and unexpected script crashes before completion.
}in the training process. To ensure consistency in evaluation, we assess all models at the end of the first epoch, after a single full pass over the novel data. For the OLMo-based score, the training process shows a sharp reward decline before completing the first full epoch. Therefore, we select an earlier checkpoint (\num{5000} steps) for evaluation, under the assumption that these \num{5000} steps still provide a meaningful degree of fine-tuning before instability occurs. 


\paragraph{1-gram BLEU Reward}

The reward starts off very low and remains low for a substantial number of steps before beginning to increase steadily. Given that this is a unigram-based metric focused on token overlap between the generated utterance and a reference, a slow and gradual increase is actually desirable, a sharp rise could lead the model to simply replicate the caregiver's utterances. For \emph{Zorro}, this model achieves the lowest score among all those evaluated, and it also ranks among the least accurate models on the \emph{dialogue minimal pairs}.  Although it is a word-based metric, no further improvements on the \emph{lexical decision} data can be reported.

\paragraph{Semantic Similarity Reward}

The reward increases gradually during training, similarly to what is observed for BLEU. However, the overall improvement across training steps is modest, and the reward values remain relatively low. On \emph{Zorro}, the model’s accuracy stays roughly at the level of \texttt{llamalogue}. Additionally, performance on the \emph{dialogue minimal pairs} shows a slight decline of a few percentage points compared to the pre-trained model. The score on the \emph{lexical decision} task is the lowest observed among all the fine-tuned models.

\paragraph{LLM-generated Reward}

Here, the reward increases during the very early phase of fine-tuning, although only by approximately $0.5$ on a scale ranging from 0 to 5. This limited growth indicates that the OLMo model used to assign the reward rarely utilized the full range of available values. In particular, scores of 0 or 5 were almost never assigned to generated utterances. Starting from around step \num{3000}, the reward begins to decline steadily. The model evaluated at checkpoint-5000 maintains a relatively strong performance on \emph{Zorro}. However, similar to the previous two PPO models, there is a decrease in accuracy on the \emph{dialogue MP} task compared to \texttt{llamalogue}. Zorro and \emph{lexical decision} scores stay roughly equivalent to the base model.

\paragraph{Teacher Confidence-based Reward}

The initial reward being around $0.2$ means the \texttt{llamalogue} reply was already above the median among the teacher’s ten candidates. At the start of fine-tuning the reward increased quickly, probably due to the initially small KL coefficient value. During fine-tuning, the reward rose to around $0.6$, meaning the fine-tuned model beats roughly eight of the teacher's alternatives. After epoch one, the reward curve had a slight dip to around $0.5$. On the \emph{lexical decision} task, the model is roughly on par with \texttt{llamalogue}, but lower on \emph{Zorro} and (slightly) \emph{dialogue MP}.

\section{Discussion and Conclusion}
\label{sec:discussion}

How can these slightly underwhelming results be explained? First, we need to emphasize that our dialogue-only models, trained on child-directed and child speech, are exposed to a smaller vocabulary \citep{snow1977talking} and simpler structures \citep{genovese2020infantdirected} than found in adult speech (although complex structures are occasionally found in CDS, they are rare, cf. \citealp{cameron-faulkner2003construction}). As the benchmarks included in BabyLM target broader lexical and syntactic variation in the input, there is a slight mismatch between our data and the evaluation data. The accuracy on lexically restricted Zorro, for example, is much higher than the one reported for BLiMP. More generally speaking, these results also align with previous findings on other models trained on CDS only (cf. \citealp{padovani2025childdirected, bunzeck2025construction}). Where our models excel is the domain of dialogue minimal pairs. There, they outperform the base model by a margin of $10\%$. While it is not overly surprising that our model masters a task that aligns $100\%$ with its pre-training goal and the shape of its data, learning dialogue coherence is still far from easy. Judging contingency and coherence without lexical overlap requires a different kind of linguistic knowledge than syntactic phenomena like island effects -- exactly the kind of knowledge our model picks up. 

With respect to the performance of our fine-tuned models, it is important to note that our results align with previous studies \citep{liu2024benchmarking, stopler2025developmentally}, which all found no significant improvements on grammatical or similar benchmarks after interaction-driven fine-tuning. Such fine-tuning with a specific, pragmatics- or communication-based goal in mind has so far only shown to improve performance on benchmarks that also test for this goal. Our DPO fine-tuning, which directly optimizes preference for correct answers, does have a positive effect on the model preferring such answers from a held-out test set. In contrast, more generalized optimization for communicatively appropriate generations with PPO does not have this effect. It remains open to further inquiry whether our scoring methods might be too abstract. After all, they are only indirectly aligned with all the different evaluation measures we want to optimize for (correct grammar, world knowledge, approximation of human reading behaviour, AoA estimation, etc.). Also, if the one, singular answer that we compare with our generation in all PPO training regimens is too distant to the generated answer (semantically, pragmatically, lexically, etc.), then the provided training signal might steer the model's weights into incorrect directions or leads to it getting stuck in local optima (exemplified by the non-monotonic reward trends).

Finally, as the differences between DPO with naturally occurring and synthetically generated answers are quite large for the dialogue MP performance, this hints towards a shortcoming of current LLMs: despite generating language that superficially resembles CDS being easy, generating authentic interactions is actually hard. For example, \citet{feng2024childdirected} generate synthetic dialogues which differ tremendously from real caretaker-child interactions -- the utterances are not fragmentary, highly verbose and complex. \citet{rasanen2024agedependent} train a CDS model from scratch, which approximates many statistical tendencies of CDS, but often generates nonsensical or ungrammatical utterances. While our model did not perform well on the general BabyLM benchmarks, a first qualitative inspection of its generative capabilities showed that it can actually continue dialogue in a plausible-looking way. Here, further experimentation with dialogue-based models is clearly needed.

\section*{Limitations}

This study has several limitations that should be acknowledged. First -- as previously discussed -- the training data is narrowly focused on child-directed and child speech, which, while intentional for our research goals, constrains the model’s lexical diversity and syntactic variety. This domain-specific bias limits generalization to broader linguistic contexts, as evidenced by weaker performance on benchmarks that target a wider range of grammatical phenomena such as BLiMP. The incorporation of adult--adult dialogue into our training regimen might be a promising direction for future research. However, our primary objective in this study was to optimize the child component’s conversational turns in dialogic interactions with caregivers, while testing if this also enhances secondary objectives like semantic relevance, common-sense reasoning, and linguistic competence. In child language development, these abilities emerge through interleaved phases/periods characterized by imitation and strong reliance (exploitation) on parental input, and others dominated by exploration of self-generated abilities and emergent capacities. Transposed to the context of a reward function guiding model competencies over time, this developmental dynamic could, for example, suggest the use of a curriculum-based reward schedule across fine-tuning steps. Such a schedule could involve intensifying the reward signal during certain stages and attenuating it during others, or alternatively optimizing different aspects of verbal production at distinct developmental phases of the model. Notably, our study did not incorporate such a curriculum in the reward design, which may have limited the effectiveness of the PPO fine-tuning. It would be interesting for future work to explore this direction and assess whether exploration/exploitation reward patterns inspired by human developmental trends could yield greater benefits for model fine-tuning. 

Furthermore, our fine-tuning phases with DPO and PPO were conducted without a previous extensive hyperparameter search. As a result, the (in-)effectiveness of our proposed reward functions and their learning dynamics remain open for further exploration. Importantly, our project is intended as a pilot study. While we put more emphasis on the comparison of and experimentation with a broad variety of studies, future work should place greater emphasis on systematically identifying the optimal hyperparameters for each reward function prior to training, thereby ensuring that observed effects can be more confidently attributed to the reward design itself rather than possibly suboptimal fine-tuning setups.

\section*{Supplementary Materials}

In addition to being conveniently available on Huggingface and GitHub, the long-term accessibility of the datasets, models, and code for the DPO and PPO experiments is ensured via a data publication on Zenodo: \url{https://doi.org/10.5281/zenodo.17253651}

\section*{Acknowledgments}

BB, MA, OM, HB and SZ acknowledge funding by the \href{https://www.dfg.de/}{Deutsche Forschungsgemeinschaft} (DFG, German Research Foundation):  \href{https://gepris.dfg.de/gepris/projekt/512393437}{CRC~1646/1~2024 -- 512393437} projects \href{https://gepris.dfg.de/gepris/projekt/537295977}{A02}, \href{https://gepris.dfg.de/gepris/projekt/537362555}{A05}, and \href{https://gepris.dfg.de/gepris/projekt/537416633}{B02}. 
FP and AB were supported by the Talent Programme of the Dutch Research Council (grant VI.Vidi.221C.009).

\bibliography{unified}

\appendix

\section{Learning trajectories across pretraining}

\begin{figure}
  \centering
    \includegraphics[width=0.95\linewidth]{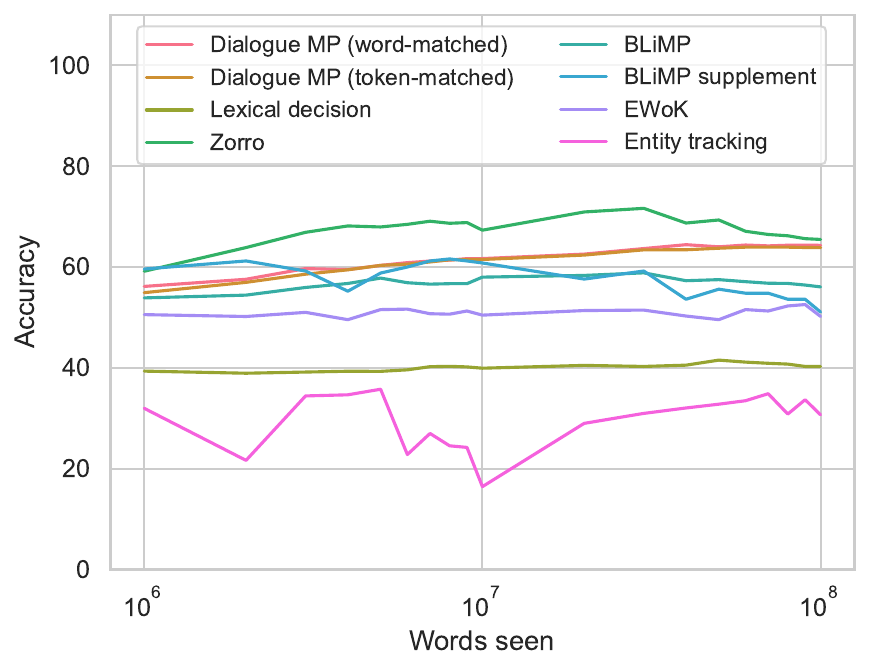}
  \caption{Learning trajectories for our base model across pre-training for 10 epochs. Note that the $x$-axis is log-scaled to make the very early training dynamics more visible.}
  \label{fig:learning_curves}
\end{figure}

To trace the learning process of \texttt{llamalogue}, we continually evaluate it during pretraining. We benchmark ten checkpoints across the first epoch (so after each 1M token set has been seen by the model once, until 10M tokens are reached) and then nine further checkpoints over the remaining nine epochs. We visualize the development of performance on eight different minimal pair sets in Figure~\ref{fig:learning_curves}.

The worst performance can be observed for the entity tracking evaluation -- performance does not stabilize at all and oscillates between $20$--$40\%$, which means that our model actively disprefers correct continuations. The same goes for the lexical decision data, where our model consistently scores around $40\%$. Performance on EWoK stays around the chance baseline as well. Interestingly, our model surpasses $60\%$ on the BLiMP supplement data around approximately 7M tokens, after which performance deteriorates again. Similarly, BLiMP performance increases slightly early on, but then also stabilizes at a low level. Accuracy scores on Zorro, the scaled-down derivative of BLiMP that only contains words also occurring in CHILDES, are generally higher and improve until the third epoch of training, after which they deteriorate again. The only stable, monotonically improving learning trajectory can be observed for our dialogue minimal pairs. This, however, is not overly surprising, as this testing paradigm aligns closely with the pretraining goal of \texttt{llamalogue}. Viewed in conjunction with our general results, these learning trajectories further corroborate the fact that the general BabyLM evaluation measures are not very suitable for our models, as the decreasing learning trajectories hint towards our models not being undertrained and because comparable studies of learning dynamics overwhelmingly report power-law like curves \citep[cf.][]{huebner2021babyberta, liu2021probing, choshen2022grammarlearning, bunzeck2024fifty, padovani2025childdirected}.

\balance
\section{DPO Datasets}
\label{sec:sample_dpo_datasets}

Table~\ref{tab:dpo-naturalistic-examples} shows a sample of sentences from the dataset we used to fine-tune the model with DPO. The appropriate and random sentences are matched in terms of token length, and both come from the distribution of sentences actually observed in CHILDES.

Table \ref{tab:dpo-synthetic-examples} was also used to fine-tune \texttt{llamalogue} with DPO. In contrast to the previous case, the appropriate sentences here are synthetic, artificially generated by Llama-3.2-3B, and their length is not matched to that of the random counterparts.

\begin{table*}[ht]
\small
\renewcommand{\arraystretch}{1.3}
\begin{tabularx}{\textwidth}{XXX}
\toprule
    \textbf{Prompt (MOT)} & \textbf{Appropriate (CHI)} & \textbf{Random (CHI)} \\  
\midrule
what is that ? & it looks like a gun . & you do it like that . \\
pull the string . & and where do they hook it ? & do you know what it was ? \\
I think they just throw it on the side . & you know what Mom ? & I get this hole . \\
what are you playing with huh toys ? & there's a dog . & there's the sports . \\
the bottom ones come off . & want to know what ? & we stole the brush . \\
can you say that ? & okay the hungry hungry caterpillar . & yeah I want that too . \\
what is it ? & a baby caterpillar ! & I'm just pretending . \\
what is it ? & I don't .. know ! & put my dress down . \\
\bottomrule
\end{tabularx}
\caption{Examples of naturalistic DPO dialogue pairs. Each row shows a caregiver's utterance (MOT) as prompt, with corresponding appropriate and random child responses (CHI) used for preference training.}
\label{tab:dpo-naturalistic-examples}
\end{table*}

\begin{table*}[ht]
\small
\renewcommand{\arraystretch}{1.3}
\begin{tabularx}{\textwidth}{XXX}
\toprule
    \textbf{Prompt (MOT)} & \textbf{Appropriate (CHI)} & \textbf{Random (CHI)} \\ 
\midrule
    what is that ? & it is a ball. & you do it like that . \\
    pull the string . & i can do it! & do you know what it was ? \\
    I think they just throw it on the side . & the garbage truck. & I get this hole . \\
    what are you playing with huh toys ? & i'm playing with this toy, it's a nice toy. & there's the sports . \\
    the bottom ones come off . & the bottom ones come off . & we stole the brush . \\
    can you say that ? & yes, mother can say that. & yeah I want that too . \\
    what is it ? & i don't know. & I'm just pretending . \\
    what is it ? & a big tree. & put my dress down . \\
\bottomrule
\end{tabularx}
\caption{Examples of synthetic DPO dialogue pairs. Each row shows a caregiver's utterance (MOT) as prompt, with corresponding appropriate child responses (CHI) generated using the Llama-3.2-3B teacher model, and random responses from naturalistic mismatches.}
\label{tab:dpo-synthetic-examples}
\end{table*}

\section{PPO Reference Child Responses}
\label{sec:ppo_ref_sent}

In Table \ref{tab:llama3b_responses}, we show a sample of 3 prompts used during fine-tuning and the 10 ground truth answers generated by Llama-3.2-3B when it is asked to simulate a child responding to a caregiver's sentence, using the prompt shown in detail in Table \ref{tab:prompt_generation}.

\begin{table*}[ht]
\centering
\footnotesize
\begin{tabular}{@{}p{0.35\textwidth}|p{0.6\textwidth}@{}}
\toprule
\textbf{Prompt (MOT)} & \textbf{Generated child-like responses (LLaMA-3B)} \\ 
\midrule
\texttt{I was wondering who told you your neck needed washing.} & \vspace{-1.7em}
\begin{enumerate}
\itemsep0em 
    \item \texttt{It was a frog.}
    \item \texttt{The neck was washed, so I asked the dog to lick it.}
    \item \texttt{No one told me.}
    \item \texttt{I don't know, mom.}
    \item \texttt{I don't know, but I think I'll wash it now.}
    \item \texttt{I don't know, I just felt it needed washing.}
    \item \texttt{I think it was you.}
    \item \texttt{It was the person in the mirror.}
    \item \texttt{Mommy told me to wash my neck.}
    \item \texttt{Mother, I was wondering who told you my neck needed washing.}
\end{enumerate} \\
\midrule
\texttt{I think they're all friends together, aren't they?} &
\vspace{-1.7em}
\begin{enumerate}
\itemsep0em 
    \item \texttt{Yes, mother, I think so.}
    \item \texttt{I think they're all friends together, aren't they?}
    \item \texttt{I think they're all friends together, aren't they?}
    \item \texttt{Yes, they are.}
    \item \texttt{Yes, mother, they are.}
    \item \texttt{Yes, I think so.}
    \item \texttt{Yes, they are, Mom.}
    \item \texttt{I think they're all friends together, aren't they?}
    \item \texttt{Yes, I think so, aren't they?}
    \item \texttt{Yes, they are friends.}
\end{enumerate} \\ 
\midrule
\texttt{You're a nice girl.} &
\vspace{-1.7em}
\begin{enumerate}
\itemsep0em 
    \item \texttt{you're a nice girl too.}
    \item \texttt{I'm sorry, I'm not a nice girl, I'm a bad girl.}
    \item \texttt{Yes, I am.}
    \item \texttt{I am a nice girl.}
    \item \texttt{thank you, mom.}
    \item \texttt{yes, I am a nice girl.}
    \item \texttt{thank you mother.}
    \item \texttt{I'm nice too.}
    \item \texttt{thank you, mom.}
    \item \texttt{yes I am.}
\end{enumerate} \\
\bottomrule
\end{tabular}
\caption{Caregiver prompts and ten possible child-like answers generated by the Llama-3.2-3B model.}
\label{tab:llama3b_responses}
\end{table*}

\section{(Super)GLUE results}
\label{apx:glue}

We report the results for the SuperGLUE tasks in Table~\ref{tab:glue-results}. Here, we can generally report that fine-tuning with DPO and PPO has only very little effect on our models' advantages for further fine-tuning. In comparison to the baseline model trained on the whole BabyLM corpus, they are generally worse base models for fine-tuning on (Super)GLUE.

\begin{table*}
\centering
\footnotesize
\begin{tabularx}{\textwidth}{Xcccccccc}
\toprule
    &  & \multicolumn{2}{c}{\textbf{DPO}} & \multicolumn{4}{c}{\textbf{PPO}} &  \\
\cmidrule(lr){3-4}\cmidrule(lr){5-8}
    \textbf{Task} & \textbf{\texttt{llamalogue}} & \textbf{Natural.} & \textbf{Synth.} & \textbf{Bleu} & \textbf{SemSim} & \textbf{LM Score} & \textbf{Conf.} & \textbf{Baseline} \\
\midrule 
    BoolQ (acc) & 64.04	& 64.04	& 64.04	& 64.04 & 64.04 &	64.04 &	64.04 & 68.38   \\
    MNLI  (acc)& 35.17	&35.17	&34.92	&34.60	&34.82	&35.23	&34.92	& 61.04\\
    MRPC (F1)& 80.95	&80.95	&80.95	&81.31	&80.95	&81.31	&80.95 & 83.61 \\ 
    QQP (F1)& 10.28	&10.28	&10.17	&5.55	&11.13	&10.37	&11.17 & 71.82 \\
    RTE (acc)& 53.24	&52.52	&53.24	&53.24	&54.68	&51.80	&53.24 & 61.15 \\
    MultiRC (acc)&57.55	&57.55	&57.55	&57.55	&57.55	&57.55	&57.55& 65.92 \\
    WSC (acc)& 61.54	&61.54	&61.54	&61.54	&61.54	&61.54	&61.54 & 63.46 \\ 
\bottomrule 
\end{tabularx}
\caption{SuperGLUE results.}
\label{tab:glue-results}
\end{table*}

\end{document}